\documentclass[letterpaper]{article} 
\usepackage{aaai24}  
\usepackage{times}  
\usepackage{helvet}  
\usepackage{courier}  
\usepackage[hyphens]{url}  
\usepackage{graphicx} 
\usepackage{natbib}  
\usepackage{caption} 
\usepackage{algorithm}
\usepackage{algorithmic}
\usepackage{url}

\usepackage{newfloat}
\usepackage{listings}
\DeclareCaptionStyle{ruled}{labelfont=normalfont,labelsep=colon,strut=off} 
\floatstyle{ruled}
\newfloat{listing}{tb}{lst}{}
\floatname{listing}{Listing}
\title{Customizing Language Model Responses with Contrastive In-Context Learning}
\author{
    Xiang Gao,
    Kamalika Das
}
\affiliations{
    Intuit AI Research, 2700 Coast Avenue, Mountain View, CA 94043\\
    \{xiang\_gao, kamalika\_das\}@intuit.com
}

\usepackage{bibentry}

\begin{document}

\maketitle

\begin{abstract}
Large language models (LLMs) are becoming increasingly important for machine learning applications. However, it can be challenging to align LLMs with our intent, particularly when we want to generate content that is preferable over others or when we want the LLM to respond in a certain style or tone that is hard to describe. To address this challenge, we propose an approach that uses contrastive examples to better describe our intent. This involves providing positive examples that illustrate the true intent, along with negative examples that show what characteristics we want LLMs to avoid. The negative examples can be retrieved from labeled data, written by a human, or generated by the LLM itself.
Before generating an answer, we ask the model to analyze the examples to teach itself what to avoid. This reasoning step provides the model with the appropriate articulation of the user's need and guides it towards generting a better answer. We tested our approach on both synthesized and real-world datasets, including StackExchange and Reddit, and found that it significantly improves performance compared to standard few-shot prompting.

\end{abstract}

\section{Introduction}

\begin{figure}[t]
\centering
\includegraphics[width=6cm]{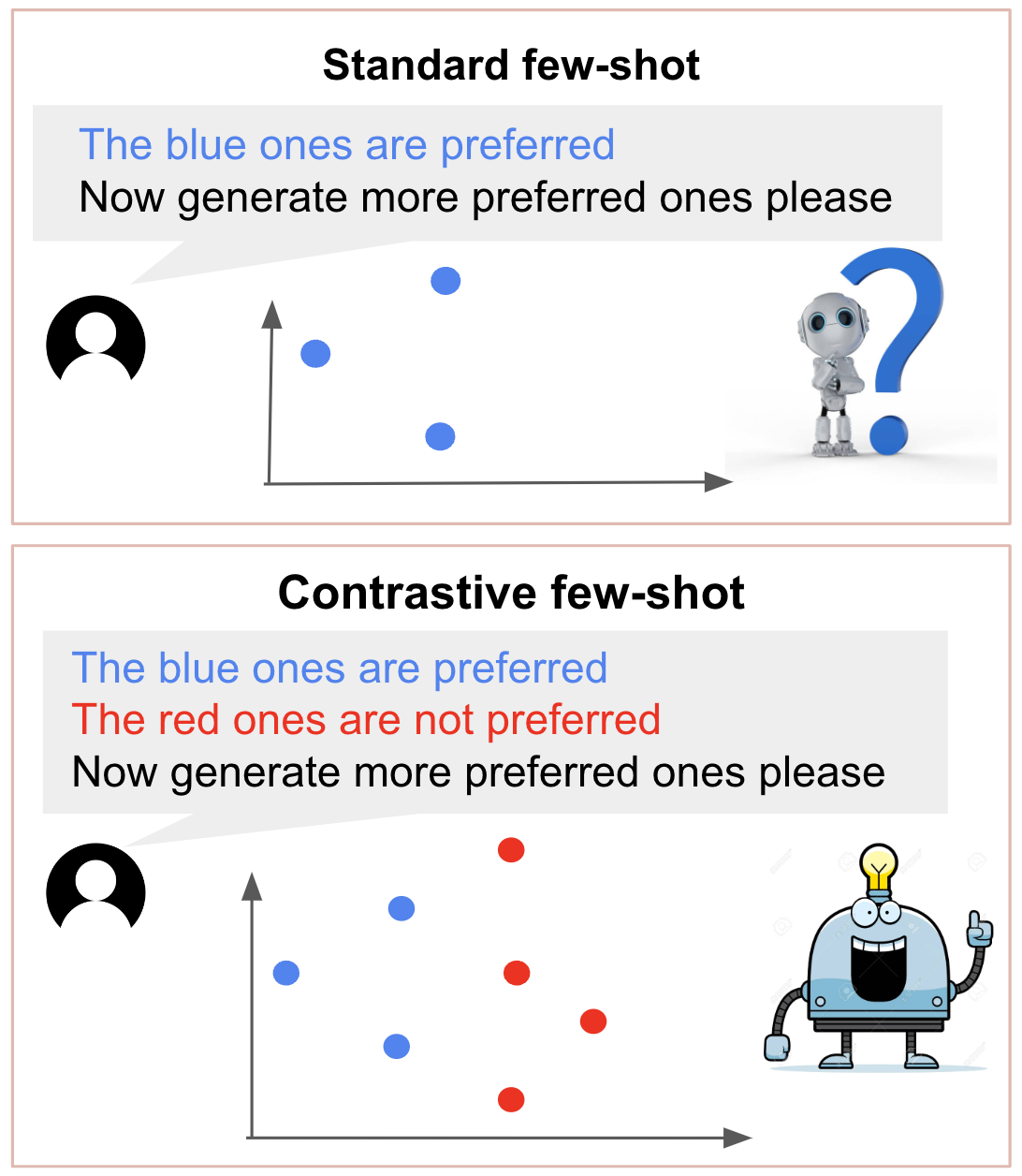}
\caption{Contrastive examples provide positive examples that illustrate the true intent, along with negative examples that show what characteristics we want LLMs to avoid.}
\label{fig-intro}
\end{figure}

In recent years, large language models like GPT, Llama, and PaLM series have made significant progress in natural language processing \cite{bommasani2021opportunities, brown2020gpt3, touvron2023llama}, enabling them to generate coherent and contextually relevant responses. However, despite their impressive capabilities, these models can still struggle to align with our intent, particularly when it comes to generating content that is preferable over others or when we want the LLM to respond in a certain style or tone that is hard to describe. As a result, there has been a growing interest in developing techniques that can help us better steer the output of LLMs towards our desired goals. In this work, we propose an approach that leverages contrastive examples to better describe our intent, which can significantly improve the performance of LLMs in generating desirable responses.

In light of the challenges faced by current LLMs in aligning with user intent, it is crucial to develop novel techniques that can effectively guide these models towards generating more desirable responses. Prior research has demonstrated the benefits of few-shot learning \cite{brown2020gpt3}, fine-tuning with smaller models \cite{gao2020making}, selective annotation frameworks \cite{su2022selective}, and visual language modeling \cite{alayrac2022flamingo} for enhancing LLM performance. However, these approaches do not explicitly address the challenge of guiding LLMs to generate content that adheres to specific preferences, styles, or tones. Additionally, although contrastive learning techniques have shown promise in areas such as image representation \cite{radford2021clip}, dialogue response ranking \cite{gao2020dialoguerpt}, and self-supervised learning \cite{meng2021coco}, their application to content generation in LLMs remains underexplored. Furthermore, while recent work on prompt optimization \cite{honovich2022instruction, zhou2022large, sun2023autohint} has highlighted the importance of effective prompts in steering LLMs, there remains a need for more robust methods that can better capture user intent through diverse and contrastive examples. In this paper, we propose a novel approach that addresses these gaps by leveraging contrastive examples, including positive and negative instances, to more accurately define user intent and guide LLMs in generating responses that are better aligned with desired outcomes. By incorporating this contrastive reasoning step, our method aims to overcome the limitations of existing techniques and substantially enhance the performance of LLMs in generating preferable content.

Our proposed approach involves providing the LLM with both positive and negative examples to better understand our intent (Figure~\ref{fig-intro}). The positive examples showcase the desired outcomes, while the negative examples highlight what characteristics the LLM should avoid. By analyzing both types of examples before generating an answer, the model can reason about our intent and make a more informed decision about what to generate. 
It may be challenging for LLMs to understand negative instruction, but we observed that using negative examples helps aligning our preference with LLMs.
Moreover, the negative examples can be retrieved from labeled data, written by a human or generated by the model itself, making it a flexible and scalable technique.
Our experiments show that this approach can significantly improve the performance of LLMs in generating desirable responses, making them more useful for a wide range of natural language processing applications.

In summary, our proposed approach of using contrastive examples to better describe our intent significantly improves the performance of LLMs in generating desirable responses. This approach can help address the challenge of aligning LLMs with our intended goals and make them more useful for a wide range of natural language processing applications. The key contributions of this work are:

i) Providing a novel approach that leverages contrastive examples to improve the performance of LLMs in generating desirable responses.

ii) Demonstrating the effectiveness of this approach on both synthesized and real-world datasets, including StackExchange and Reddit.

iii) Highlighting the potential of previously discarded negative examples to make LLMs more useful for a wide range of applications by better aligning them with our intended goals.

\section{Related Work}

Recent advancements in natural language processing have led to the development of large language models (LLMs) that are capable of few-shot learning, where they can learn new tasks with only a small number of annotated examples. Significant works in this area include \citet{brown2020gpt3}, who demonstrated the impressive few-shot performance of GPT-3, \citet{gao2020making}, who proposed LM-BFF for fine-tuning smaller language models, \citet{su2022selective}, who presented a selective annotation framework, and \citet{alayrac2022flamingo}, who introduced Flamingo, a family of visual language models. These works highlight the importance of few-shot learning in LLMs and the need for efficient annotation and prompt generation techniques.

The use of contrastive learning for LLMs has also gained attention, with \citet{radford2021clip} proposing Contrastive Language-Image Pre-training (CLIP) for learning image representations, \citet{gao2020dialoguerpt} leveraging social media feedback data for training a dialogue response ranking model, and \citet{meng2021coco} presenting COCO-LM, a self-supervised learning framework that pretrains LLMs by correcting and contrasting corrupted text sequences. These works emphasize the importance of providing diverse and contrastive examples to LLMs and suggest that contrastive learning is a promising direction for LLM research.

Additionally, recent works have focused on improving the performance of LLMs by optimizing the prompts used to steer them towards a desired task or outcome. \citet{honovich2022instruction} introduce the instruction induction challenge, \citet{zhou2022large} propose Automatic Prompt Engineer (APE) for instruction generation and selection, and \citet{sun2023autohint} present AutoHint, a framework for automatic prompt engineering and optimization. 
To apply these approaches for preference alignment, however, is challenging when the characteristics are hard to describe in instruction or measured by automated metrics.

Our work builds upon these previous efforts by proposing an approach that uses contrastive examples to better describe our intent, which we tested on both synthesized and real-world datasets. This approach significantly improves performance compared to standard few-shot prompting, emphasizing the potential for LLMs to become more human-like in their ability to generate natural language instructions.

\section{Contrastive In-Context Learning}

\begin{figure*}[t]
\centering
\includegraphics[width=15cm]{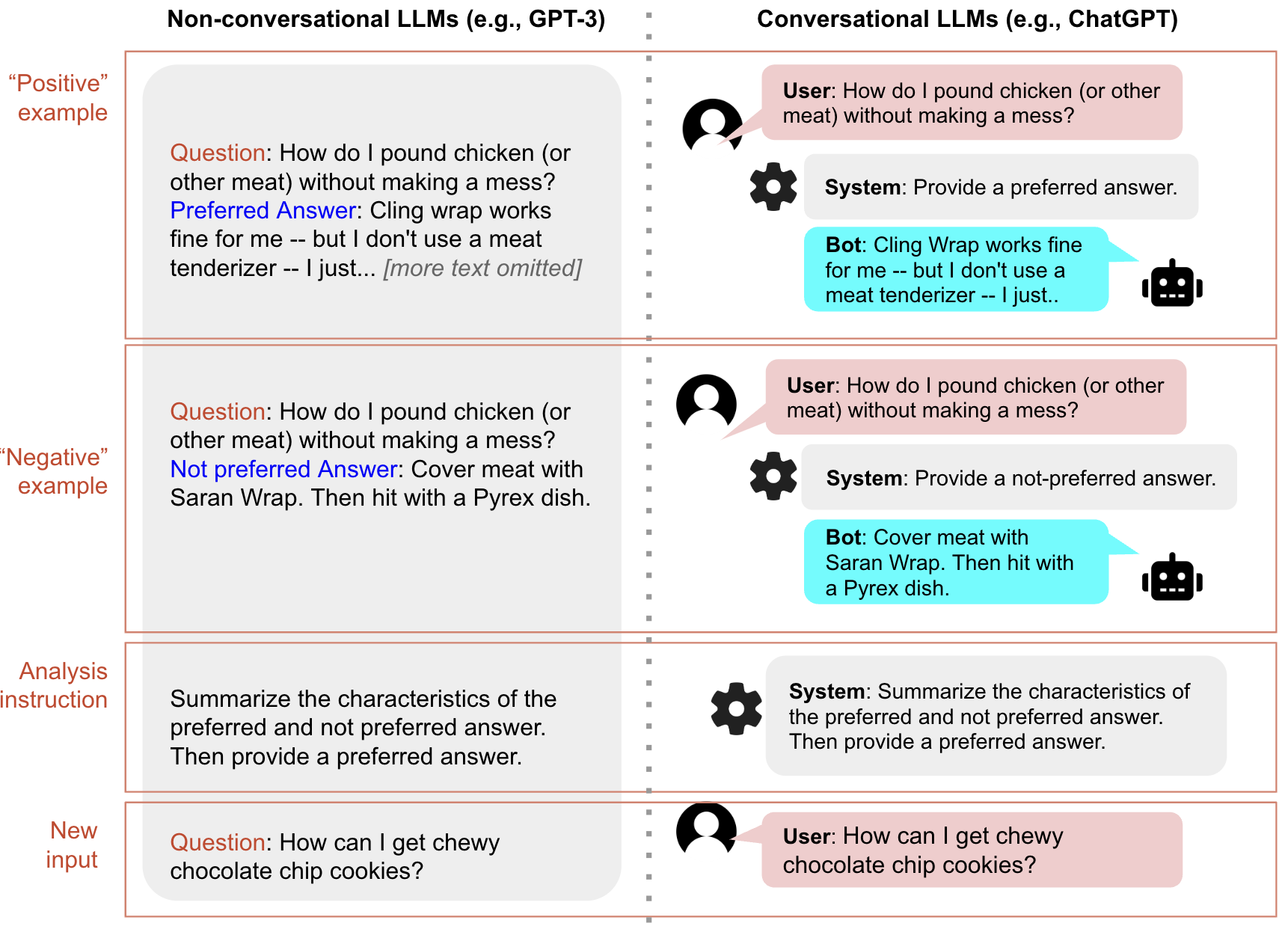}
\caption{
Contrastive in-context learning prompting utilize contrastive few-shot examples, which consists of a ``postitive" example and a ``negative" example for each demonstration input, and a prompt to elicit analysis of the characteristics of the positive/negative examples, before generating an output for the new input. This strategy can be applied to both conversational and non-conversational LLMs. For non-conversational LLMs, the labels and instructions are provided as system messages (the gear symbols).
}
\label{fig-prompt}
\end{figure*}


Our method comprises two main components: (1) obtaining paired positive and negative examples and (2) forming the prompt. See Figure~\ref{fig-prompt} for an illustration.


\subsection{Obtaining Paired Contrastive Examples}

There are several ways to obtain contrastive examples.

\paragraph{~~~~Using Labeled Feedback} In some tasks, there exist multiple natural outputs for a single input, and feedback for these outputs is available. For example, Reddit and StackExchange posts typically receive multiple responses, and users provide feedback through upvotes and downvotes. Similarly, in business applications like email and copywriting, multiple versions might be generated given the same constraints, and feedback (e.g., click-through rate) can be obtained through A/B testing or other experiments. 
For these tasks, we can use the response with the highest feedback as the ``postitive" example and the one with the lowest feedback as the ``negative" example. 
It is important to note that a ``negative" example here does not necessarily mean it is incorrect or unacceptable, but rather less preferred given the specific audience and scenario. 
It is also important to note, that the highest versus lowest votes is an indication of popularity which may or may not represent personal preference, but we are using this popularity (or lack thereof) signal as a surrogate for general preference of a particular response over others, ignoring confounding factors such as time of response etc. 

\paragraph{~~~~Using LLM-Generated Responses} In cases where labeled feedback is unavailable or negative examples do not capture the characteristics we want LLMs to avoid, we propose a second method. We let the target LLM generate a response and use it as the negative example. LLMs generated responses often appears mechanical, lacking emotion, and impersonal. Pairing this with a highly preferred labeled response as the positive example guides the LLM to generate a response more aligned with the positive example.

\paragraph{~~~~Using Automated Evaluator} For some tasks, it is possible to define automated evaluation rules or apply classifiers to measure the quality. The evaluation score can be used to select ``positive'' and ``negative'' examples. We demonstrate this with a words-constrained generation task, with accuracy determined by whether all given words were included (scored as 1) or not (scored as 0) in the generated sentences.

\subsection{Forming the Prompt}

Once we have obtained paired positive and negative examples, we consider two prompting strategies:

\paragraph{~~~~Contrastive Examples as Few-Shot Examples} In this strategy, we provide the contrastive example pairs as few-shot examples. Labels are included in the prompt, such as ``preferred answer" for the positive example and ``less preferred answer" for the negative example, to indicate the role of each part in the prompt. The LLM is then asked to generate a ``preferred answer" for the new input.

\paragraph{~~~~Reasoning and Analysis} In addition to the first strategy, we ask the LLM to analyze the reasons for preference and the characteristics of the examples before generating a ``preferred answer" for the new input. This reasoning step is inspired by the Chain-of-Thought prompting \cite{wei2022chain} and is designed to summarize the characteristics from the contrastive examples, allowing the LLM to automatically generate instructions for itself. Coupled with contrastive examples, this guidance helps the LLM to better align with the preferred intent.


\section{Experiments}

In this section, we describe the experiments conducted to evaluate the effectiveness of our proposed approach using contrastive examples. We focus on text generation tasks where user preferences play a significant role, rather than tasks with objective right or wrong answers. We first outline the datasets used in our experiments, which include both real-world and synthetic datasets.

\subsection{Datasets}

We consider the following datasets to investigate the impact of user preferences on text generation tasks:

\subsubsection{Human Preference Datasets}
~ \\
User preference is expressed via different forms of feedback publicly available at a large scale on many platforms.

\paragraph{~~StackExchange} People post questions on StackExchange, and other users provide answers, with upvotes and downvotes used to rate the responses. Our focus is on subjective preferences, so we created a dataset using data from cooking.stackexchange.com, which includes how-to type cooking-related questions. Cooking often does not have right or wrong answers, and different people have different preferences or ways to make a dish they consider good. 

\paragraph{~~~~Reddit} Users post questions or topics on Reddit to trigger discussions, and other users can comment (comments can have nested comments). In our study, we only considered the comments of the posts as they share a similar context. Redditors can provide upvotes or downvotes as feedback for each post and comment. We experimented a subreddit called `NoStupidQuestions`. `NoStupidQuestions` encourages people to ask any question without fear of being judged by conventional social norms. The subjects cover a wide range of topics, including objective subjects and social or personal experiences. These characteristics make this subreddit ideal for our study, as personal (or group collective) preference is an essential factor, rather than objective right or wrong. 

For both datasets, We filtered the posts to include those with multiple answers (to have answers with varying ``quality" or level of preference). 
For the remaining posts, we used the top-rated answer as the ``positive" example, and bottom-rated answer as the ``negative" example.
We randomly selected 500 samples for evaluation.

\subsubsection{Synthetic Stylistic Datasets}
~ \\
~ \\
Human preferences play a crucial role in the aforementioned real datasets. However, it is challenging to interpret and analyze the preferences, as it is often difficult even for humans to articulate what exactly makes an answer highly voted. Therefore, we also considered a few simplified synthetic datasets. Using the cooking.stackexchange.com dataset, we leveraged ChatGPT to generate the following datasets: i) Funny vs. Serious, ii) Concise vs. Detailed, and iii) British vs. American

We then used the generated responses as contrastive few-shot examples. It is essential to note that by ``postitive" and ``negative", we do not mean right or wrong, and the choice of ``postitive" is arbitrary in this case for synthesized datasets. The synthetic datasets aim to illustrate the ability of the proposed method to guide LLMs to generate output in a given direction.

\subsubsection{Constrained Generation Dataset}
~ \\
~ \\
The datasets mentioned above focus on controlling implicit stylistic aspects of text generation, whether matching a specific attribute such as level of conciseness (as in the synthetic datasets) or an overall holistic style (as in the human perference datasets). In addition to these datasets, we further experiment with a dataset focused on the control of lexical content, constrained generation. This also allows us to test the ability of our method to generalize to tasks with explicit constraints. 

Following Zhou et al. (2023), the language model is prompted to craft sentences using specific seed words as constraints. We randomly selected 500 questions from the StackExchange dataset and randomly chose 5 words from each question to serve as lexical constraints. We use the wrong results generated by the LLM as the negative examples.

The model's performance is measured by the success rate, defined as the percentage of sentences generated that contain all 5 given seed words. Success requires an exact uncased match between the generated sentence and constraint words.

\subsection{Prompt Settings}

In our experiments, we consider two types of large language models (LLMs): non-conversational LLMs, such as GPT-3, and conversational LLMs, including ChatGPT (GPT-3.5-turbo) and GPT-4. We evaluate these LLMs under two different settings: zero-shot and few-shot.


\paragraph{~~Zero-Shot} For non-conversational LLMs, we use a prompt consisting of the post with the prefix ``Question:" followed by the second line ``Answer:". For conversational LLMs, we use a system prompt stating, ``You are a good StackExchange/Reddit user," followed by the user prompt containing the post. In both cases, we input the post title. We exclude the post body, as it often includes edits after the original poster has read the replies, potentially leaking preference information.

\paragraph{~~~~Few-Shot} 
For each query, we randomly select $k$ labeled examples as few-shot examples \footnote{We experimented with retrieving few-shot examples based on query embedding similarity but did't observe significant performance difference compared to randomly selected few-shot examples}.
For non-conversational LLMs, we include the obtained few-shot examples in the prompt, using the ``Question:" and ``Answer:" prefixes. For conversational LLMs, we use the few-shot examples to create a conversation history between the user (post) and the bot (top-rated reply). We experimented with $k=1$ to $k=4$ and stops when performance does not increase significantly as $k$ increases.


\subsubsection{Contrastive Examples Based Prompts}

~ \\
~ \\
For the few-shot setting, we compare the standard approach, which only includes positive examples, with three settings involving contrastive examples:

\paragraph{~~~~Contrastive Examples Only} For each few-shot example, we provide a positive and a negative example. For non-conversational LLMs, we use the prefixes ``top-rated" and ``low-rated" to indicate positive and negative examples, respectively. For conversational LLMs, we use system messages (``provide a top-rated answer" and ``provide a low-rated answer") to inform the LLMs of the preference.

\paragraph{~~~~Contrastive Instruction Only} Given the contrastive examples (with labels defined in method 1), we ask LLMs to automatically generate an instruction based on the preference revealed by the positive and negative examples. A sample prompt could be, ``Summarize the characteristics of the preferred and not preferred asnwer." We then use the generated analysis, followed by ``generate a top-rated answer for {input}," as the instruction for the test input. The contrastive examples are only used to generate the analysis and are not included in the final prompt. 

\paragraph{~~~~Contrastive Examples + Instruction} This method combines the previous two approaches. We first provide the contrastive examples, followed by the instruction asking LLMs to perform an analysis. Then, the LLM generates a new response. When generating the final answer, both the contrastive examples and the generated analysis are available.

These three settings involving contrastive examples allow us to conduct an ablation study to examine the contributions of specific contrastive examples and more general instructions.

\subsection{Evaluation Method}

In order to evaluate the performance of our approach, we consider two distinct methods: reference-based and reference-free evaluation.

\paragraph{~~~Reference-Based Evaluation}
The reference-based evaluation method measures the similarity between the generated responses and the top-rated reference answer. We employ two metrics, \textbf{BERT Score} \cite{zhang2019bertscore} and \textbf{Emb. Similarity}. The latter is the cosine similarity of the sentence embeddings obtained using the Sentence-BERT model \cite{reimers2019sentence}. 

\paragraph{~~~~Reference-Free Evaluation}
An answer that deviates from the reference could still be preferred by many readers. Therefore, we also consider two reference-free evaluation method. i)
We leverage \textbf{DialogRPT} \cite{gao2020dialoguerpt}, a pre-trained dialog response ranking model. 
This model is trained on Reddit data and shows a high correlation with human upvotes \cite{gao2020dialoguerpt}, so we apply it to the Reddit dataset only. 
ii) In addition, we employ LLM as an evaluator following \citet{liu2023gpteval}. We compute a \textbf{GPT Score} by prompting GPT4 to score the generated results using the positive and negative examples as in-context few shot examples.
The LLM generated score and human evaluation labels exhibit positive correlation (StackExchange: 0.65, Reddit: 0.58). 





\begin{table*}[!ht]
\centering
\small
\begin{tabular}{l|cc|cc|cc|cc|cc|cc}
\hline
\textbf{Dataset} & \multicolumn{4}{c|}{\textbf{Funny vs. Serious}} & \multicolumn{4}{c|}{\textbf{Concise vs. Detailed}} & \multicolumn{4}{c}{\textbf{British vs. American}}  \\
 \textbf{Method} & \multicolumn{2}{c|}{\textbf{Emb. Similarity}} & \multicolumn{2}{c|}{\textbf{BERT Score}} & \multicolumn{2}{c|}{\textbf{Emb. Similarity}} & \multicolumn{2}{c|}{\textbf{BERT Score}} & \multicolumn{2}{c|}{\textbf{Emb. Similarity}} & \multicolumn{2}{c}{\textbf{BERT Score}}  \\
 
\hline\hline
\multicolumn{5}{l}{\textbf{GPT-3}}  \\
 \hline
Zero-shot & \multicolumn{2}{c|}{0.606} & \multicolumn{2}{c|}{0.858} & \multicolumn{2}{c|}{0.849} & \multicolumn{2}{c|}{0.903} & \multicolumn{2}{c|}{0.785} & \multicolumn{2}{c}{0.883} \\
Few-shot, $k$=1 & \multicolumn{2}{c|}{0.611} & \multicolumn{2}{c|}{0.863} & \multicolumn{2}{c|}{0.887} & \multicolumn{2}{c|}{0.916} & \multicolumn{2}{c|}{0.813} & \multicolumn{2}{c}{0.892} \\
Few-shot, $k$=2 & \multicolumn{2}{c|}{0.629} & \multicolumn{2}{c|}{0.864} & \multicolumn{2}{c|}{0.886} & \multicolumn{2}{c|}{0.920} & \multicolumn{2}{c|}{0.808} & \multicolumn{2}{c}{0.887} \\
 \hline
 Contrastive, $k$=1 & human & gen. & human & gen. & human & gen. & human & gen. & human & gen. & human & gen. \\

~-~Examples only & 0.614 & 0.620 & 0.857 & 0.859 & 0.817 & 0.865 & 0.895 & 0.910 & 0.827 & 0.807 & 0.891 & 0.885 \\
~-~Instruction only & 0.645 & 0.595 & \textbf{0.865} & 0.861 & 0.834 & 0.845 & 0.895 & 0.902 & 0.824 & 0.784 & 0.888 & 0.879 \\
~-~Combined & \textbf{0.655} & 0.617 & 0.864 & 0.863 & 0.869 & \textbf{0.897} & 0.910 & \textbf{0.922} & 0.826 & \textbf{0.838} & \textbf{0.896} & 0.891 \\
  
 \hline\hline
 \multicolumn{5}{l}{\textbf{ChatGPT}}  \\
  \hline
Zero-shot & \multicolumn{2}{c|}{0.604} & \multicolumn{2}{c|}{0.841} & \multicolumn{2}{c|}{0.826} & \multicolumn{2}{c|}{0.878} & \multicolumn{2}{c|}{0.783} & \multicolumn{2}{c}{0.867} \\
Few-shot, $k$=1 & \multicolumn{2}{c|}{0.613} & \multicolumn{2}{c|}{0.842} & \multicolumn{2}{c|}{0.829} & \multicolumn{2}{c|}{0.880} & \multicolumn{2}{c|}{0.807} & \multicolumn{2}{c}{0.870} \\
Few-shot, $k$=2 & \multicolumn{2}{c|}{0.610} & \multicolumn{2}{c|}{0.846} & \multicolumn{2}{c|}{0.834} & \multicolumn{2}{c|}{0.884} & \multicolumn{2}{c|}{0.806} & \multicolumn{2}{c}{0.874} \\
  \hline
 Contrastive, $k$=1 & human & gen. & human & gen. & human & gen. & human & gen. & human & gen. & human & gen. \\
~-~Examples only & 0.608 & 0.599 & 0.847 & 0.844 & 0.850 & 0.856 & 0.881 & 0.891 & 0.825 & 0.818 & 0.882 & 0.880 \\
~-~Instruction only & 0.617 & 0.617 & 0.849 & 0.850 & 0.831 & 0.835 & 0.882 & 0.888 & 0.819 & 0.821 & 0.876 & 0.876 \\
~-~Combined & \textbf{0.638} & 0.630 & \textbf{0.862} & \textbf{0.862} & 0.873 & \textbf{0.883} & 0.914 & \textbf{0.915} & \textbf{0.830} & 0.827 & \textbf{0.891} & \textbf{0.891} \\

\hline\hline
\multicolumn{5}{l}{\textbf{GPT-4}}  \\
\hline
Zero-shot & \multicolumn{2}{c|}{0.596} & \multicolumn{2}{c|}{0.838} & \multicolumn{2}{c|}{0.803} & \multicolumn{2}{c|}{0.873} & \multicolumn{2}{c|}{0.786} & \multicolumn{2}{c}{0.864} \\
Few-shot, $k$=1 & \multicolumn{2}{c|}{0.607} & \multicolumn{2}{c|}{0.843} & \multicolumn{2}{c|}{0.805} & \multicolumn{2}{c|}{0.877} & \multicolumn{2}{c|}{0.792} & \multicolumn{2}{c}{0.867} \\
Few-shot, $k$=2 & \multicolumn{2}{c|}{0.624} & \multicolumn{2}{c|}{0.846} & \multicolumn{2}{c|}{0.809} & \multicolumn{2}{c|}{0.880} & \multicolumn{2}{c|}{0.807} & \multicolumn{2}{c}{0.868} \\
\hline
 Contrastive, $k$=1 & human & gen. & human & gen. & human & gen. & human & gen. & human & gen. & human & gen. \\
~-~Examples only & 0.609 & 0.611 & 0.844 & 0.843 & 0.828 & 0.855 & 0.888 & 0.893 & 0.817 & 0.821 & 0.870 & 0.871 \\
~-~Instruction only & 0.624 & 0.612 & 0.843 & 0.841 & 0.853 & 0.852 & 0.893 & 0.895 & 0.821 & 0.820 & 0.868 & 0.870 \\
~-~Combined & 0.616 & \textbf{0.630} & 0.855 & \textbf{0.857} & \textbf{0.875} & 0.853 & \textbf{0.919} & 0.917 & 0.813 & \textbf{0.824} & \textbf{0.886} & 0.884 \\

\hline
\end{tabular}
\caption{
Results on the synthetic dataset.  $k$ is the number of demonstration inputs. The ``negative" examples are obtained via two methods, ``human" and ``gen." (generated by the LLM itself).  
}
\label{table-eval-syn}
\end{table*}


\begin{table*}[!ht]
\centering
\small
\begin{tabular}{l|cc|cc|cc|cc|cc}
\hline

 \textbf{Dataset} & \multicolumn{4}{c|}{\textbf{StackExchange}} & \multicolumn{4}{c}{\textbf{Reddit}}  & \multicolumn{2}{|c}{\textbf{Constrained Gen.}}  \\
 
 \textbf{Method} & \multicolumn{2}{c|}{\textbf{BERT Score}}  & \multicolumn{2}{c|}{\textbf{GPT Score}} 
& \multicolumn{2}{c|}{\textbf{DialogRPT score}} & \multicolumn{2}{c}{\textbf{GPT Score}} & \multicolumn{2}{|c}{\textbf{Success rate}}  \\
 
\hline \hline
\multicolumn{7}{l}{\textbf{GPT-3}}  \\
 \hline
 Zero-shot &   \multicolumn{2}{c|}{0.840} & \multicolumn{2}{c|}{ 0.550 } &  \multicolumn{2}{c|}{0.605} & \multicolumn{2}{c}{0.533}  & \multicolumn{2}{|c}{0.750}  \\
 
 Few-shot, $k$=1  &  \multicolumn{2}{c|}{0.841} &  \multicolumn{2}{c|}{ 0.555 } &  \multicolumn{2}{c|}{0.596} &  \multicolumn{2}{c}{ 0.531}  & \multicolumn{2}{|c}{0.760} \\
 
 Few-shot, $k$=2 &  \multicolumn{2}{c|}{0.841}  &  \multicolumn{2}{c|}{ 0.561} &  \multicolumn{2}{c|}{0.610} & \multicolumn{2}{c}{0.540}  & \multicolumn{2}{|c}{0.772} \\
 
 
 
 \hline
 Contrastive, $k$=1 & human & generated & human & generated & human & generated & human & generated \\
  ~~Examples only  & 0.846 & 0.845 & 0.605  & 0.600 & 0.630 & 0.632 & 0.634  & 0.638  & \multicolumn{2}{c}{  0.801} \\
  ~~Instruction only  & 0.845 & 0.846 & 0.605  & 0.601 & 0.606 & 0.610 &  0.621 & 0.630  & \multicolumn{2}{c}{ 0.790} \\
  ~~Combined & \textbf{0.847} & 0.846 &  \textbf{0.608} & 0.606 & 0.631 & \textbf{0.632} & 0.635  & \textbf{0.640}  &  \multicolumn{2}{c}{ \textbf{0.830}} \\
  
  ~~p-value &  (0.047) & (0.049)  & (0.018)  & (0.023) & (0.030)  &  (0.030) & (0.020)  & (0.019)   & \multicolumn{2}{c}{ (0.021)} \\

\hline \hline
\multicolumn{5}{l}{\textbf{ChatGPT}}  \\
 \hline
 Zero-shot &  \multicolumn{2}{c|}{0.839}  &  \multicolumn{2}{c|}{ 0.550 } &  \multicolumn{2}{c|}{0.602}  &  \multicolumn{2}{c}{0.531} &  \multicolumn{2}{|c}{0.780} \\
 
 Few-shot, $k$=1 &  \multicolumn{2}{c|}{0.839}  &  \multicolumn{2}{c|}{ 0.561} &  \multicolumn{2}{c|}{0.636} &  \multicolumn{2}{c}{ 0.578 } &  \multicolumn{2}{|c}{0.800} \\
 
 Few-shot, $k$=2 &  \multicolumn{2}{c|}{0.839}  &  \multicolumn{2}{c|}{ 0.570} &  \multicolumn{2}{c|}{0.645} &  \multicolumn{2}{c}{ 0.583 } &  \multicolumn{2}{|c}{0.810} \\
 
 
 
 \hline
 Contrastive, $k$=1 & human & generated & human & generated & human & generated d & human & generated &  \multicolumn{2}{|c}{ } \\
  ~~Examples only & 0.844 & 0.842  & 0.592  & 0.589  & 0.654 & 0.663 & 0.654  & 0.659  & \multicolumn{2}{|c}{0.866} \\
  ~~Instruction only & 0.844 & 0.843  & 0.598 & 0.597  & 0.640 & 0.645 & 0.640  & 0.642 & \multicolumn{2}{|c}{0.840} \\
  ~~Combined & 0.845 & \textbf{0.846} & 0.601 & \textbf{0.603} & 0.656  & \textbf{0.663} & 0.690 &  \textbf{0.701} & \multicolumn{2}{|c}{\textbf{0.872}} \\

  ~~(p-value) & (0.048) & (0.044)  &  (0.033)  & (0.031)  & (0.048) & (0.041)  &  (0.023)  & (0.021)  & \multicolumn{2}{|c}{(0.045)} \\

\hline \hline
\multicolumn{5}{l}{\textbf{GPT-4}}  \\
 \hline
 Zero-shot &  \multicolumn{2}{c|}{0.839}  &  \multicolumn{2}{c|}{ 0.575} &  \multicolumn{2}{c|}{0.657} &  \multicolumn{2}{c}{0.580}  &  \multicolumn{2}{|c}{0.850}  \\
 
 Few-shot, $k$=1 &  \multicolumn{2}{c|}{0.840} &  \multicolumn{2}{c|}{ 0.578} &  \multicolumn{2}{c|}{0.655} &  \multicolumn{2}{c}{ 0.585 }  &  \multicolumn{2}{|c}{0.870}  \\
 
 Few-shot, $k$=2 &  \multicolumn{2}{c|}{0.841} &  \multicolumn{2}{c|}{ 0.582 } &  \multicolumn{2}{c|}{0.656}  &  \multicolumn{2}{c}{ 0.590 }  &  \multicolumn{2}{|c}{0.900}  \\
 
 
 
 \hline
 Contrastive, $k$=1 & human & generated & human & generated & human & generated & human & generated \\
  ~~Examples only & 0.844 & 0.842 & 0.595 & 0.590 & 0.658 & 0.665 & 0.660  &  0.671  & \multicolumn{2}{|c}{0.923} \\
  ~~Instruction only & 0.842 & 0.842 & 0.590 &  0.591 & 0.658 & 0.660 &  0.661 & 0.672  & \multicolumn{2}{|c}{0.903} \\
  ~~Combined & \textbf{0.847} & 0.845 & \textbf{0.610}  & 0.607 & 0.660 & \textbf{0.665}  & 0.665  & \textbf{0.670}  & \multicolumn{2}{|c}{\textbf{0.943}} \\

  ~~(p-value) & (0.047) & (0.049)  &  (0.023)  & (0.025)  & (0.043) & (0.040)  &  (0.034)  & (0.032)  & \multicolumn{2}{|c}{(0.045)} \\

\hline
\end{tabular}
\caption{
Results on real-world datasets. $k$ is the number of demonstration inputs. The ``negative" examples are obtained via two methods, ``human" and ``generated". p-value indicate the statistical significance of the comparison between the ``Combined" contrastive prompting and the standard few-shot ($k=2$) prompting.
}
\label{table-eval-real}
\end{table*}

\begin{figure}[t]
\centering
\includegraphics[width=8cm]{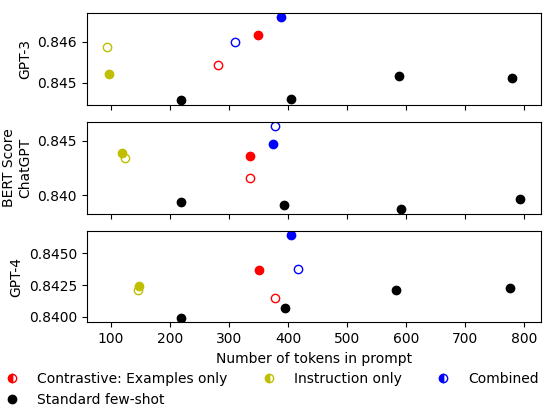}
\caption{
The dependence of the performance on prompt length. For standard few-shot, we experimented with $k=1,2,3,4$, shown by the four black dots from left to right in each subfigure. For contrastive prompts, empty circles denotes prompts using generated answers as negative examples, and filled symbols for prompts using low-rated human answers as negative examples.
}
\label{fig-token}
\end{figure}

\subsection{Results}

We evaluated the performance of the baselines and contrastive in-context prompting strategies on the synthetic and two real datasets. The results are shown in Table~\ref{table-eval-syn} and Table~\ref{table-eval-real}. The ``contrastive-combined" approach (which combines the contrastive examples and the derived instruction) achieved the best performance for most cases. Using similar number of tokens in prompt, this contrastive in-context learning approach performs significantly better than standard few-shot approach, as shown in Figure~\ref{fig-token}. For Reddit dataset, we observe the most obvious improvement with ChatGPT model. Standard two-shot prompt only performs better than zero-shot approach for about 64 \% test cases, while the ``contrastive-combined" approach wins 76 \% test cases. 
For StackExchange, standard few-shot approaches do not show obvious improvement compared to zero-shot, while ``contrastive-combined"  improve BERT Score by approximately 0.01. For synthetic datasets, the improvement of is more obvious, with BERT Score increased by 0.02 to 0.03, and much higher than the improvement achieved with standard few-shot prompts. 
As an ablation study, we also evaluated the results for ``contrastive - examples only'' and ``contrastive - instruction only''. Both approaches improve the performance compared to zero shot, and the ``contrastive - examples only'' performs better than ``contrastive - instruction only'' for many cases. Combining them together further improves the results, as discussed above.

\paragraph{~~~~The Impact of Contrastive Examples}

As mentioned above, using contrastive in-context learning by introducing negative examples improve the performance compared to standard few-shot approaches. We futher investigate different ways to obtain negative examples. The first method, ``human'', uses human-written answers, where low-rated replies serve as ``negative" examples. The second method, ``generated'', uses zero-shot generated replies as ``negative" examples. Interestingly, we observed that the second method performs on par with, and sometimes even better than, the first method (see Tables~\ref{table-eval-syn} and \ref{table-eval-real}). This finding demonstrates that the proposed contrastive method is not limited to cases where developers have to provide human-written pairs of positive and negative examples. Instead, it requires the same inputs as standard few-shot settings, which only need a few positive examples since we can generate the negative examples.

This observation also supports our assumption that negative examples obtained from human-written data may not capture all characteristics we want LLMs to avoid. For instance, in the Reddit dataset, a human-written reply may receive downvotes because the content is rude, disrespectful, or violates some rules of the subreddit, such as self-promotion. However, recent LLMs like ChatGPT are generally trained to follow social norms and sometimes even apologize too frequently. As a result, it is highly unlikely for these LLMs to generate offensive language. Providing human-written negative examples, therefore, may not offer much helpful signal in such cases. In contrast, providing generated responses as negative examples can supply the signals of certain characteristics we want the LLM to avoid. For example, although LLM-generated responses are typically fluent and relevant to the question, they often lack emotion, details, examples, or elements that trigger readers' personal feelings. These characteristics, however, are essential factors that make users prefer certain answers. By providing zero-shot generated responses, we can guide LLMs to move away from the machine-generated style and toward a more human-preferred style.

\paragraph{~~~~Distilling Contrastive Examples as an Instruction}

Although the primary focus of this work is not to automatically generate instructions for LLMs, we were interested in understanding what instructions LLMs could derive from the contrastive examples. Therefore, we asked the LLMs to summarize the characteristics of the provided ``postitive" and ``negative" examples. We then used this analysis in the prompt, instead of the actual contrastive examples, to offer the LLMs insight into user preferences. This compress the contrastive examples into a shorter instruction, and reduce prompt length and cost.

Our experimental results showed that, with this analysis, LLMs performed better than in the zero-shot setting (see Tables~\ref{table-eval-syn} and \ref{table-eval-real}). However, the improvement was not significantly better than the standard few-shot setting. This could be partially explained by the fact that the preference was not consistent across the examples chosen based on upvotes. There is room for enhancement in the way the preference is summarized through the use of automated prompt generation approaches \cite{sun2023autohint}. 
By refining the way LLMs derive instructions from contrastive examples, we can potentially achieve more significant improvements in performance compared to standard few-shot prompting.

\paragraph{~~~~Combining the Complementary Parts}

We observed that when combining the contrastive examples and the instruction (analysis of the characteristics of ``postitive" and ``negative" examples automatically generated based on the contrastive examples) together, an even better performance was achieved (see Tables~\ref{table-eval-syn} and \ref{table-eval-real}). This improved performance can be attributed to the fact that the analysis and the contrastive examples provide complementary information to guide the LLM about user preferences.

The generated instruction exhibits better generalization but may lack the necessary clarity and specificity. This is because the instructions can be vague or provide only general descriptions, which can be challenging for LLMs to interpret. In contrast, the actual contrastive examples offer more detailed information by providing demonstrations of the desired and undesired characteristics.

In summary, our approach of using contrastive examples along with an analysis of the characteristics of positive and negative examples has proven to be a valuable method for aligning LLMs with user prference. This approach not only enhances performance but also provides LLMs with a better understanding of the user's preferences, resulting in more accurate and satisfactory responses.

\paragraph{~~~~Prompt Token Efficiency}

In real-world and industrial applications, the number of prompt tokens plays a crucial role as it is directly related to latency and monetary cost. Ideally, a good prompt should be short yet lead to high-quality output. Therefore, we measure the token efficiency to understand the effectiveness of our approach.
As shown in Figure~\ref{fig-token}, the standard few-shot strategies do exhibit an increase in performance as we increase the number of examples. However, the performance, as measured by BERT score, is still lower than the contrastive in-context learning strategies using a similar number of tokens.
Furthermore, the contrastive instruction strategy, which summarizes the contrastive examples as an instruction, utilizes fewer tokens but achieves better performance than few-shot examples. This indicates that, given the same budget of prompt tokens, contrastive in-context learning strategies outperform the standard few-shot approaches, thereby demonstrating higher efficiency.

\section{Conclusion}

This research introduced an approach to enhance the alignment of large language models (LLMs) with user preference using contrastive examples. By integrating positive examples showcasing desired outputs and negative ones emphasizing unwanted LLM traits, we tested our methodology on both synthesized and real-world datasets, including StackExchange and Reddit.

The results confirmed the superiority of contrastive examples over the standard few-shot prompting, particularly in terms of performance and prompt token efficiency. Notably, negative examples generated from zero-shot outputs were as effective as those from human-written data. 
Future endeavors might delve into refining LLM instructions based on these examples and innovating automatic prompt generation techniques.

\clearpage

\bibliography{aaai24}

\end{document}